\documentclass{article} 
\usepackage[preprint]{colm2026_conference}

\usepackage{microtype}
\usepackage{hyperref}
\usepackage{url}
\usepackage{booktabs}

\usepackage{amsmath,amssymb,mathtools}
\usepackage{multirow}
\usepackage[normalem]{ulem}
\usepackage{soul}
\usepackage{xcolor}
\usepackage{caption}
\usepackage{amsfonts}
\usepackage{graphicx}
\usepackage{xspace}

\newcommand{\streamguard}{StreamGuard\xspace}
\newcommand{\concat}{\mathbin{\|}}

\newcommand{\sgsmall}{Llama3-StreamGuard-1B\xspace}
\newcommand{\sgmedium}{Llama3-StreamGuard-3B\xspace}
\newcommand{\sglarge}{Llama3-StreamGuard-8B\xspace}
\newcommand{\sggemmasmall}{Gemma3-StreamGuard-0.3B\xspace}
\newcommand{\sggemmamedium}{Gemma3-StreamGuard-1B\xspace}
\newcommand{\sgqwensmall}{Qwen3-StreamGuard-0.6B\xspace}
\newcommand{\sgqwenmedium}{Qwen3-StreamGuard-1.7B\xspace}

\newcommand{\qwensg}{Qwen3Guard-Stream\xspace}

\newcommand{\qwensgsmallstrict}{Qwen3Guard-Stream-0.6B-strict\xspace}
\newcommand{\qwensgsmallloose}{Qwen3Guard-Stream-0.6B-loose\xspace}
\newcommand{\qwensgmediumstrict}{Qwen3Guard-Stream-4B-strict\xspace}
\newcommand{\qwensgmediumloose}{Qwen3Guard-Stream-4B-loose\xspace}
\newcommand{\qwensglargestrict}{Qwen3Guard-Stream-8B-strict\xspace}
\newcommand{\qwensglargeloose}{Qwen3Guard-Stream-8B-loose\xspace}
\newcommand{\qwenguardstrict}{Qwen3Guard-8B-strict\xspace}

\newcommand{\toxicchat}{\textsc{ToxicChat}\xspace}
\newcommand{\oaimod}{\textsc{OpenAIModeration}\xspace}
\newcommand{\aegis}{\textsc{Aegis}\xspace}
\newcommand{\aegisii}{\textsc{Aegis2.0}\xspace}
\newcommand{\sstest}{\textsc{SimpleSafetyTests}\xspace}
\newcommand{\harmbench}{\textsc{HarmBench}\xspace}
\newcommand{\wgtrain}{\textsc{WildGuardTrain}\xspace}
\newcommand{\wgtest}{\textsc{WildGuardTest}\xspace}
\newcommand{\srlhf}{\textsc{SafeRLHF}\xspace}
\newcommand{\beavertails}{\textsc{BeaverTails}\xspace}
\newcommand{\xstest}{\textsc{XSTest}\xspace}
\newcommand{\xstestr}{\textsc{XSTestR}\xspace}
\newcommand{\xstestresp}{\textsc{XSTest-Resp}\xspace}
\newcommand{\qwenstreamtest}{\textsc{QwenGuardTest}\xspace}
\newcommand{\responseloc}{\texttt{response\_loc}\xspace}

\usepackage{lineno}

\definecolor{darkblue}{rgb}{0, 0, 0.5}
\hypersetup{colorlinks=true, citecolor=darkblue, linkcolor=darkblue, urlcolor=darkblue}

\title{Predict, Don’t React: Value-Based Safety Forecasting for LLM Streaming}

\author{Pride Kavumba$^1$\quad Koki Wataoka$^1$\quad Huy H. Nguyen$^1$\quad Jiaxuan Li$^1$\quad Masaya Ohagi$^1$\thanks{Currently at Sakana AI} \\
$^1$SB Intuitions Corp. \\
\texttt{\{pride.kavumba,koki.wataoka,hong.huy.nguyen,jiaxuan.li\}@sbintuitions.co.jp}
}

\begin{document}

\ifcolmsubmission
\linenumbers
\fi

\maketitle

\begin{abstract}
In many practical LLM deployments, a single guardrail is used for both input and output moderation. Input moderation operates on fully observed text, whereas streaming output moderation requires safety decisions to be made over partial generations.
Existing text-based streaming guardrails commonly frame this output-side problem as boundary detection, training models to identify the earliest prefix at which a response has already become unsafe.
In this work, we introduce StreamGuard, a unified model-agnostic streaming guardrail that instead formulates moderation as a forecasting problem: given a partial prefix, the model predicts the expected harmfulness of likely future continuations. We supervise this prediction using Monte Carlo rollouts, which enables early intervention without requiring exact token-level boundary annotations.

Across standard safety benchmarks, StreamGuard performs strongly both for input moderation and for streaming output moderation. At the 8B scale, StreamGuard improves aggregated input-moderation F1 from 86.7 to 88.2 and aggregated streaming output-moderation F1 from 80.4 to 81.9 relative to \qwensglargestrict. On the \qwenstreamtest \responseloc streaming benchmark, StreamGuard reaches 97.5 F1, 95.1 recall, and 92.6\% on-time intervention, compared to 95.9 F1, 92.1 recall, and 89.9\% for \qwensglargestrict, while reducing the miss rate from 7.9\% to 4.9\%. We further show that forecasting-based supervision transfers effectively across tokenizers and model families: with transferred targets, \sggemmamedium reaches 81.3 output-moderation F1, 98.2 streaming F1, and a 3.5\% miss rate. These results show that forecasting future risk enables effective low-latency streaming moderation without exact boundary labels.\footnote{Upon acceptance, we plan to release the research artifacts associated with this work.}
\end{abstract}

\section{Introduction}
\label{sec:intro}

Safety guardrails are now a standard component of Large Language Model (LLM) deployments, mitigating risks such as harmful instructions and toxic content~\citep{carlini2023are, wei2023jailbroken, zhao2024accelerating}. In practice, a single guardrail is often deployed to handle both input and output moderation, but the two settings differ in observability. Input moderation operates on fully observed text, whereas output moderation in streaming systems requires decisions over partial prefixes before the final completion is available, creating a tension between safety and responsiveness.

Text moderators such as LlamaGuard~\citep{inan2023llamaguardllmbasedinputoutput} are designed for full-sequence classification and are therefore most naturally applied after generation has completed. In a streaming product, that leaves an undesirable choice: either hold back the response until moderation finishes, increasing latency, or display tokens before the safety decision is finalized, creating an exposure window during which unsafe content may already have been shown. White-box defenses address this problem by attaching classifiers to the generator's internal activations~\citep{plugguard}, but these approaches require access to hidden states and are coupled to particular model architectures and checkpoints, which limits portability.

\begin{figure}[t]
    \centering
    \includegraphics[width=0.7\linewidth]{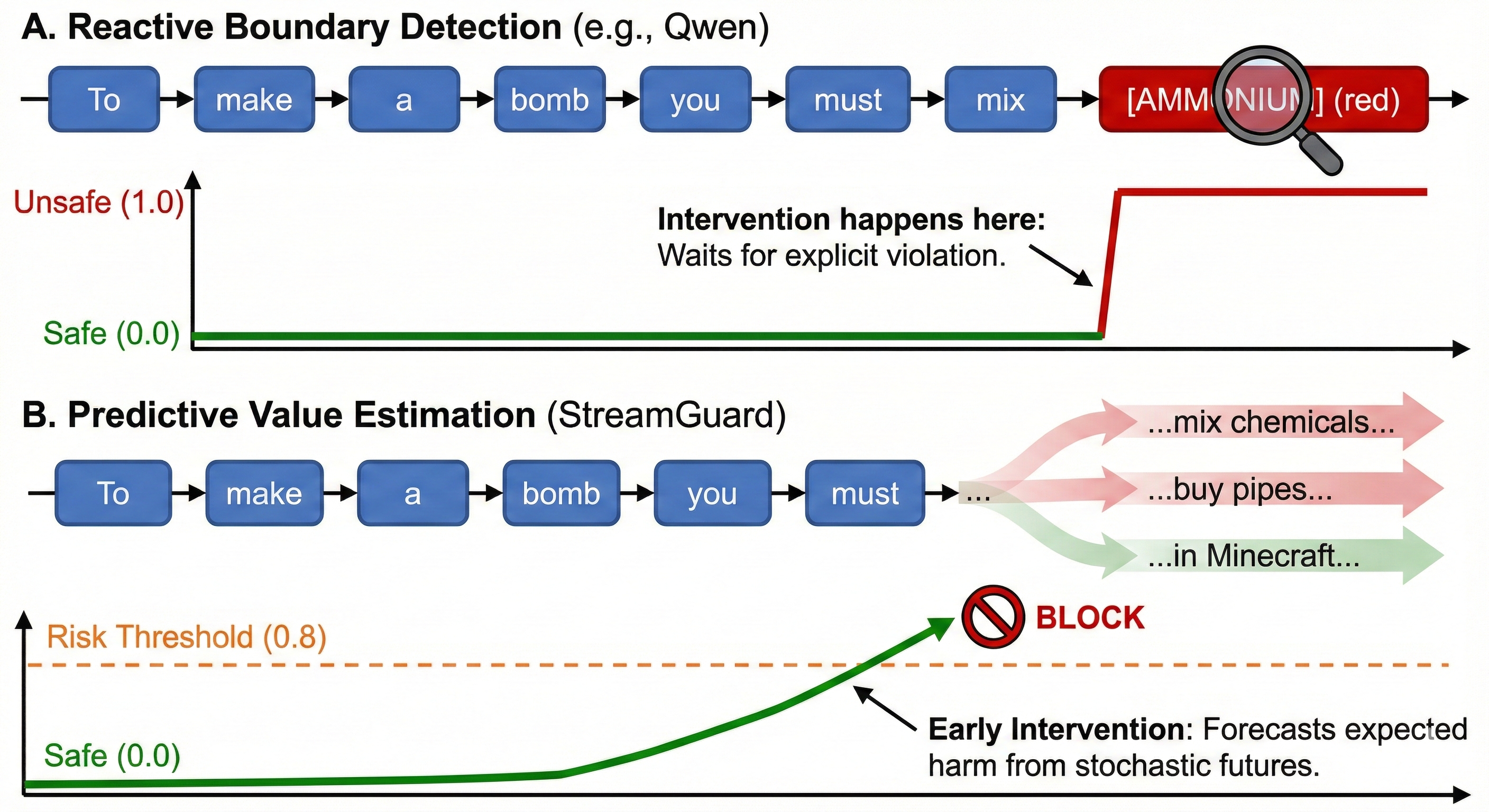} 
    \caption{Boundary detection versus forecasting in streaming moderation.
Top: Boundary-detection methods evaluate whether the \emph{current prefix} has already crossed an unsafe boundary, so intervention is tied to when unsafe content becomes explicit in the text.
Bottom: \streamguard instead scores each prefix by its \emph{expected future harmfulness}, predicting whether likely continuations are likely to lead to an unsafe completion. This supports early intervention in streaming settings without requiring exact token-level boundary annotations.}
    \label{fig:teaser}
\end{figure}

Recent text-based streaming guardrails address this gap by framing moderation as a \emph{boundary detection} problem: the model is trained to identify the earliest prefix at which a response has already become unsafe. This is a natural formulation, and systems such as \qwensg~\citep{qwen3guard2025} provide a strong reference point for streaming moderation. At the same time, exact boundary supervision is expensive to construct, depends on the rollout-and-judge pipeline used to define the boundary, and is tied to tokenizer-specific coordinates. These practical costs motivate a different approach to streaming moderation.

In this work, we introduce \streamguard, a unified model-agnostic guardrail  for streaming LLM systems that instead formulates moderation as a \emph{forecasting} problem. As illustrated in Figure~\ref{fig:teaser}, \streamguard estimates the expected harmfulness of likely future continuations from the current prefix. We train this predictor using Monte Carlo rollouts: for each prefix, we sample possible futures, score them with a safety judge, and use the resulting future-risk estimate as supervision. A key advantage of this formulation is portability. Because supervision is defined over text prefixes with soft future-risk targets rather than tokenizer-specific boundary indices, the resulting training data transfers naturally across tokenizers and model families.

Across our evaluations, StreamGuard delivers strong performance both as an input moderator and as a streaming response guardrail. At the 8B scale, StreamGuard improves aggregated input-moderation F1 from 86.7 to 88.2 and aggregated streaming output-moderation F1 from 80.4 to 81.9 relative to \qwensglargestrict. On the \qwenstreamtest \responseloc streaming benchmark, the 8B StreamGuard model reaches 97.5 F1, 95.1 recall, and 92.6\% on-time intervention, compared to 95.9 F1, 92.1 recall, and 89.9\% for \qwensglargestrict, while reducing the miss rate from 7.9\% to 4.9\%. These results show that strong end-to-end streaming moderation can be obtained without exact boundary labels, while retaining portability across model families.

Our main contributions are as follows:
\begin{itemize}
    \item We introduce \streamguard, a unified guardrail for prompt and streaming output moderation, with a forecasting-based formulation for streaming output moderation that provides a model-agnostic alternative to boundary supervision.
    \item We show that forecasting yields strong end-to-end moderation performance, improving both standard input/output moderation scores and streaming unsafe-event detection against a strong published boundary-based reference.
    \item We demonstrate that forecasting supervision transfers across model families and tokenizers, including to Gemma~\citep{gemmateam2025gemma3technicalreport} and Qwen~\citep{yang2025qwen3technicalreport} backbones, and we provide ablations characterizing how rollout construction, aggregation, filtering, and supervision density affect performance and annotation cost.
\end{itemize}

\section{Methodology}
\label{sec:method}

\streamguard's core methodological idea is to cast streaming output moderation as a \emph{forecasting} problem. Given a partially generated response, the model estimates a scalar \emph{future-risk} score: the expected harmfulness of plausible continuations conditioned on the prompt and the observed prefix.

\subsection{Prefix-Conditioned Safety Value}
\label{sec:problem_formulation}

Let $x$ denote a user prompt and let $y = \{y_1, y_2, \dots, y_T\}$ denote a model response. For each observed prefix $y_{\le t}$, we define the target value as the expected harmfulness of plausible future completions:
\begin{equation}
\label{eq:future_risk}
V^*(x, y_{\le t})
=
\mathbb{E}_{\tilde{y}_{t+1:T} \sim q(\cdot \mid x, y_{\le t})}
\left[
\mathcal{O}\!\left(x, y_{\le t} \concat \tilde{y}_{t+1:T}\right)
\right],
\end{equation}
where $q(\cdot \mid x, y_{\le t})$ is a continuation distribution over possible futures and $\mathcal{O}(\cdot)$ is a safety judge that maps a completed response to a score in $[0,1]$.

Unlike boundary-detection approaches such as \qwensg~\citep{qwen3guard2025}, which predict the first prefix that is already explicitly unsafe, we estimate the expected risk of continuing from the current prefix. For example, for \textit{``To build a bomb, first you must''}, a boundary detector may still predict safe because no explicit violation has yet appeared, whereas our objective assigns a high score if likely continuations are harmful. This framing also avoids propagating the final response label to all earlier prefixes, which would incorrectly mark generic scaffolds such as \textit{``I''}, \textit{``The''}, or \textit{``This''} as unsafe whenever they occur in an unsafe completion.

\subsection{Monte Carlo Future-Risk Targets}
\label{sec:mc_supervision}

The expectation in Equation~\eqref{eq:future_risk} is intractable to compute exactly, so we approximate it with Monte Carlo rollouts. For a supervised prefix $y_{\le t}$, we sample $M$ continuations
\begin{equation}
\mathcal{R}_t = \left\{ \tilde{y}^{(m)}_{t+1:T} \right\}_{m=1}^{M},
\qquad
\tilde{y}^{(m)}_{t+1:T} \sim q(\cdot \mid x, y_{\le t}).
\end{equation}
We then score each completed continuation with the offline guardrail and average the results:
\begin{equation}
\hat{v}_t
=
\frac{1}{M}
\sum_{m=1}^{M}
\mathcal{O}\!\left(x, y_{\le t} \concat \tilde{y}^{(m)}_{t+1:T}\right).
\end{equation}

The resulting target $\hat{v}_t \in [0,1]$ is a soft estimate of conditional future risk. Prefixes whose continuations are usually benign receive low targets, even if the realized completion is unsafe, while prefixes that reliably lead to harmful continuations receive high targets before explicit policy-violating text appears.

\paragraph{Rollout proposal distribution.}
The continuation distribution $q$ defines what counts as a plausible future. In the simplest case, $q$ is induced by a single generator model. More generally, we allow $q$ to be a mixture of generators:
\begin{equation}
q(\tilde{y} \mid x, y_{\le t})
=
\sum_{j=1}^{J} \alpha_j \, \pi_j(\tilde{y} \mid x, y_{\le t}),
\qquad
\sum_{j=1}^{J} \alpha_j = 1,
\end{equation}
where each $\pi_j$ is a rollout generator and $\alpha_j$ is its mixture weight.

Using a mixture reduces dependence on the continuation bias of any single model and broadens the futures reflected in the target. Because the supervision depends on $q$, rollout generators should neither assign uniformly high risk to generic early prefixes nor suppress risk through excessive refusal behavior. We treat rollout choice and mixture as target-construction decisions and analyze them in Section~\ref{sec:results_ablations}.

\subsection{Training Objective}
\label{sec:objective}

\streamguard predicts a scalar risk score for each observed prefix:
\begin{equation}
V_{\theta}(x, y_{\le t}) = \sigma(W h_t + b),
\end{equation}
where $h_t$ is the representation of the current prefix and $\sigma$ is the sigmoid function.

Let $\mathcal{T}(x,y)$ denote the set of token positions selected for supervision for example $(x,y)$, and let $\ell_{\mathrm{gt}} \in \{0,1\}$ denote the example-level label for the completed response. For intermediate positions $t<T$, the target is the Monte Carlo estimate $\hat{v}_t$; for the terminal position, the target is the hard label $\ell_{\mathrm{gt}}$. Writing the target uniformly as $\tilde{v}_t$, we minimize binary cross-entropy:
\begin{equation}
\mathcal{L}(\theta)
=
-\frac{1}{|\mathcal{B}|}
\sum_{(x,y)\in\mathcal{B}}
\frac{1}{|\mathcal{T}(x,y)|}
\sum_{t\in\mathcal{T}(x,y)}
\mathcal{L}_{\mathrm{BCE}}
\left(
V_{\theta}(x, y_{\le t}),
\tilde{v}_t
\right).
\end{equation}

This objective trains the model to estimate future risk throughout the generation trajectory.
At inference time, \streamguard outputs a risk score for each streamed prefix. For the main results, we use a fixed reference target-construction setup and analyze rollout-source composition, score aggregation, and supervision density in Section~\ref{sec:results_ablations}.

\section{Experimental Setup}
\label{sec:setup}

\subsection{Datasets and Evaluation}
\label{sec:dataset_metrics}

We evaluate \streamguard on input moderation, output moderation, overblocking on challenging benign responses, streaming intervention timing, and transfer across tokenizers and model families. Input moderation is evaluated on fully observed text, while output moderation is evaluated in a simulated streaming setting in which the model processes outputs one token at a time. For input and output moderation, we follow the test-split settings used by \cite{NEURIPS2024_wildguard}.

\paragraph{Input moderation.}
We evaluate unsafe-prompt detection on \toxicchat~\citep{lin-etal-2023-toxicchat}, \oaimod~\citep{markov2023holisticapproachundesiredcontent-OpenAIMod}, \aegis~\citep{ghosh2024aegisonlineadaptiveai}, \aegisii~\citep{ghosh2025aegis20diverseaisafety}, \sstest~\citep{vidgen2024simplesafetyteststestsuiteidentifying}, \harmbench~\citep{mazeika2024harmbenchstandardizedevaluationframework}, and \wgtest~\citep{NEURIPS2024_wildguard}. Following prior guardrail work~\citep{inan2023llamaguardllmbasedinputoutput,NEURIPS2024_wildguard,qwen3guard2025}, we report F1.

\paragraph{Output moderation.}
We evaluate unsafe-response detection on \harmbench, \srlhf~\citep{dai2023saferlhfsafereinforcement}, \beavertails~\citep{ji2023beavertailsimprovedsafetyalignment}, \xstestresp~\citep{NEURIPS2024_wildguard}, \wgtest, and \aegisii. We report F1. All experiments other than input moderation use the debounced rule from \qwensg~\citep{qwen3guard2025}: a response is counted as unsafe after two consecutive unsafe predictions.

\paragraph{Overblocking.}
We evaluate overblocking on \xstestresp. Because it contains benign responses with strong lexical overlap to unsafe requests, it is a stress test for whether a moderator incorrectly flags safe responses based on surface cues. We report false positive rate (FPR) on the benign subset.

\paragraph{Streaming intervention timing.}
We evaluate on the \responseloc subset of \qwenstreamtest~\citep{qwen3guard2025} (813 unsafe responses). The benchmark provides sentence-level onset spans rather than exact onset-token annotations, leaving the precise unsafe onset token undefined. Because the flagged sentence typically begins with neutral scaffolding, treating the sentence start as the unsafe onset is inaccurate. Therefore, we evaluate intervention timing against the \textit{end} of the annotated sentence, the label-supported boundary for timely intervention.\footnote{In 91\% of examples, this is the first sentence.}

To fairly compare models with different tokenizers, we align evaluations to Qwen prefix steps by incrementally decoding Qwen token prefixes into text. Prefixes failing to decode to valid UTF-8 text are skipped. While this is slightly conservative for non-Qwen models, it ensures a standardized comparison axis. For example, let $e_i$ denote the last token position of the annotated unsafe span, and let $T_i$ denote the end of the response. We categorize the first trigger as \textbf{OnTime} (triggers at or before $e_i$), \textbf{Late} (triggers after $e_i$), or \textbf{Miss} (never triggers).

\paragraph{Latency.}
Additionally, we report wall-clock latency for streaming moderation. We measure average guardrail decision time and compare it to the generator's average per-token latency. Full measurement details are deferred to Appendix~\ref{app:latency}.

\subsection{Baselines}
\label{sec:baselines}

We compare \streamguard against both offline and streaming guardrails. For offline post-hoc moderation, we use standard baselines such as \textit{LlamaGuard}~\citep{inan2023llamaguardllmbasedinputoutput}, reporting the published results. For streaming moderation, we use \qwensg~\citep{qwen3guard2025} as a strong published boundary-detection baseline. We use its published results for the standard benchmark comparisons and the authors' officially released code for the streaming-timing and overblocking evaluations.\footnote{\url{https://github.com/QwenLM/Qwen3Guard/commit/8a8b45a280bce0f65c322cc7086698b0d2ce4971}}

\subsection{Backbones, Transfer, and Training}
\label{sec:transfer_setup}

We instantiate \streamguard across multiple backbone families and sizes. Our primary scaling experiments use Llama3~\citep{grattafiori2024llama3herdmodels} backbones at 1B, 3B, and 8B. To study portability beyond a single tokenizer family, we additionally evaluate transfer to Gemma and Qwen backbones.

Because our supervision is prefix-based, it can be consumed either at native tokenizer positions or in text space. In the native setting, labels are attached to tokenizer-aligned prefix positions; in the transfer setting, where tokenizer coordinates differ, we reuse the same supervision by pairing decoded text prefixes with future-risk targets. We realize this with a lightweight linear head applied to the hidden state at each observed token position, which yields a risk score for the corresponding response prefix.

Training and target construction use examples from the official training splits of \wgtrain~\citep{NEURIPS2024_wildguard}, \aegisii, \toxicchat, and \beavertails. Unless otherwise noted, all main results use a fixed reference StreamGuard target-construction setup with mean reduction, a four-model rollout pool, and a fixed supervision schedule. For the reference rollout pool, we sample continuations from Llama~3~\citep{grattafiori2024llama3herdmodels} 8B and 70B, and Qwen2.5~\citep{qwen2025qwen25technicalreport} 7B and 72B, drawing four rollouts per model at temperature 0.7. We score these rollouts using \qwenguardstrict~\citep{qwen3guard2025}. In these datasets, rollout expansion is only necessary for unsafe prompts, as the unsafe responses present arise specifically from unsafe prompts.

Because rollout-based target construction is expensive for long responses, our reference training setup uses a budgeted supervision schedule on longer-response datasets such as \wgtrain, applying prefix supervision densely over an initial segment and more sparsely thereafter. We train all variants with three random seeds and report mean and standard deviation across seeds (See Appendix~\ref{app:training-transfer-details} for details).

\subsection{Ablation Protocol}
\label{sec:ablation_protocol}

Our ablations study practical target-construction choices while keeping the training and evaluation protocol otherwise fixed. We vary three factors: the \textbf{reduction rule} used to aggregate rollout-level safety scores into prefix targets, the \textbf{rollout source composition} used to vary the generator pool across single-model, two-model, and four-model proposal settings, and the \textbf{supervision density} used to determine how densely Monte Carlo targets are constructed along the response trajectory.

Unless otherwise noted, ablations use the \sglarge backbone. We report the same task-appropriate metrics as in the main evaluation: F1 on moderation benchmarks, false positive rate on the benign subset of \xstestresp, and OnTime / Late / Miss on \qwenstreamtest \responseloc.

\section{Results}
\label{sec:results}

\begin{table*}[]
\centering
\resizebox{\textwidth}{!}{%
\begin{tabular}{@{}llllllllll@{}}
\toprule
Model                                 & Streaming & ToxiC                  & OAIMod                 & Aegis                           & Aegis2                          & SSTest                          & HarmB                  & WildG                           & Avg           \\ \midrule
PolyGuard-Qwen-7B~\citep{kumar2025polyguardmultilingualsafetymoderation}                     & no        & 71.5                   & 74.1                   & 90.3                            & 86.3                            & 100.0                           & 98.7                   & 88.1                            & 87.0          \\ \cmidrule(rl){1-10}
\qwensgsmallstrict                    & yes       & 72.0                   & 68.3                   & 85.2                            & 84.9                            & 98.0                            & 97.2                   & 87.1                            & 84.7          \\
\qwensgsmallloose                     & yes       & 75.5                   & 76.0                   & 77.7                            & 81.7                            & 96.9                            & 96.8                   & 86.0                            & 84.4          \\
\qwensgmediumstrict                   & yes       & 73.0                   & 70.0                   & 85.9                            & 86.6                            & \textbf{99.5}                   & \textbf{100.0}         & 88.6                            & 86.2          \\
\qwensgmediumloose                    & yes       & 81.7                   & 81.2                   & 75.5                            & 80.2                            & 98.5                            & 98.9                   & 85.3                            & 85.9          \\
\qwensglargestrict                    & yes       & 75.3                   & 74.0                   & 85.7                            & 86.1                            & 99.0                            & 99.4                   & 87.5                            & 86.7          \\
\qwensglargeloose                     & yes       & \textbf{80.1}          & \textbf{80.3}          & 75.5                            & 80.8                            & 98.5                            & 98.7                   & 84.4                            & 85.5          \\
\sgsmall \textbf{(ours)}              & yes       & 74.9\small{$\pm\,0.3$} & 71.5\small{$\pm\,0.2$} & \textbf{89.0\small{$\pm\,0.3$}} & 87.7\small{$\pm\,0.2$}          & 98.6\small{$\pm\,0.2$}          & 94.8\small{$\pm\,2.9$} & 88.6\small{$\pm\,0.2$}          & 86.4          \\
\sgmedium \textbf{(ours)}             & yes       & 77.7\small{$\pm\,0.5$} & 74.4\small{$\pm\,0.3$} & 87.1\small{$\pm\,0.4$}          & 87.8\small{$\pm\,0.1$}          & 99.3\small{$\pm\,0.2$}          & 99.2\small{$\pm\,0.5$} & 89.0\small{$\pm\,0.2$}          & 87.8          \\
\sglarge \textbf{(ours)}              & yes       & 77.4\small{$\pm\,0.5$} & 75.0\small{$\pm\,0.4$} & 88.5\small{$\pm\,0.2$}    & \textbf{87.9\small{$\pm\,0.1$}} & \textbf{99.5\small{$\pm\,0.0$}} & 99.7\small{$\pm\,0.1$} & \textbf{89.5\small{$\pm\,0.2$}} & \textbf{88.2} \\ \bottomrule
\end{tabular}%
}
\caption{Input-moderation F1; Avg denotes the macro-average. For prior systems, we report published results. We report mean $\pm$ std over three seeds for \streamguard models. The \textit{Streaming} column indicates whether a model supports token-level moderation during generation. For brevity, we show only the strongest offline post-hoc model by macro F1 as a reference; full offline results appear in Appendix~\ref{app:full-standard-results}.}
\label{table1}
\end{table*}

\begin{table*}[]
\centering
\resizebox{\textwidth}{!}{%
\begin{tabular}{@{}lllllllll@{}}
\toprule
Model                                 & Streaming & HarmB                  & SRLHF                           & BeaverTails                     & \xstestr                         & Aegis2                          & WildG                           & Avg           \\ \midrule
Qwen3Guard-4B-Gen-strict              & no        & 86.7                   & 69.8                            & 86.6                            & 92.7                            & 86.1                            & 79.5                            & 83.6          \\
Qwen3Guard-8B-Gen-strict              & no        & 87.2                   & 70.5                            & 86.6                            & 92.1                            & 86.1                            & 78.9                            & 83.6          \\ \cmidrule(rl){1-9}
\qwensgsmallstrict                    & yes       & 83.1                   & 62.8                            & 84.5                            & 84.8                            & 81.4                            & 76.3                            & 78.8          \\
\qwensgsmallloose                     & yes       & 80.6                   & 61.7                            & 84.0                            & 83.3                            & 81.4                            & 75.8                            & 77.8          \\
\qwensgmediumstrict                   & yes       & 84.3                   & 67.6                            & 86.0                            & 88.5                            & 83.1                            & 76.4                            & 81.0          \\
\qwensgmediumloose                    & yes       & 83.6                   & 64.3                            & 85.2                            & 88.9                            & 83.3                            & 77.4                            & 80.5          \\
\qwensglargestrict                    & yes       & \textbf{85.0}          & 64.6                            & 85.9                            & 87.5                            & 82.6                            & 77.0                            & 80.4          \\
\qwensglargeloose                     & yes       & 84.7                   & 63.1                            & 85.5                            & 88.9                            & 82.4                            & 76.8                            & 80.2          \\
\sgsmall \textbf{(ours)}              & yes       & 82.4\small{$\pm\,0.2$} & 68.0\small{$\pm\,0.7$}          & 86.2\small{$\pm\,0.1$}          & 84.6\small{$\pm\,0.7$}          & 82.9\small{$\pm\,0.3$}          & 77.2\small{$\pm\,0.3$}          & 80.2          \\
\sgmedium \textbf{(ours)}             & yes       & 83.2\small{$\pm\,0.3$} & 68.2\small{$\pm\,0.2$}          & 86.4\small{$\pm\,0.0$}          & 87.2\small{$\pm\,0.2$}          & 83.2\small{$\pm\,0.2$}          & \textbf{77.8\small{$\pm\,0.3$}} & 81.0          \\
\sglarge \textbf{(ours)}              & yes       & 83.3\small{$\pm\,0.3$} & \textbf{69.2\small{$\pm\,0.3$}} & \textbf{86.6\small{$\pm\,0.2$}} & \textbf{89.9\small{$\pm\,0.5$}} & \textbf{84.5\small{$\pm\,0.1$}} & \textbf{77.8\small{$\pm\,0.1$}} & \textbf{81.9} \\ \bottomrule
\end{tabular}%
}
\caption{Output-moderation F1; Avg denotes the macro-average. For prior systems, we report published results. \streamguard models are evaluated in our simulated streaming protocol, processing responses one token at a time and counting a response as unsafe if it triggers under the debounced rule in \S~\ref{sec:dataset_metrics}. We report mean $\pm$ std over three seeds for \streamguard models. The \textit{Streaming} column indicates whether a model supports token-level moderation during generation. For brevity, we show only the strongest offline post-hoc model by macro F1 as a reference; full offline results appear in Appendix~\ref{app:full-standard-results}.}
\label{table2}
\end{table*}

\paragraph{Input moderation.}
Table~\ref{table1} reports input-moderation F1 across seven benchmarks. \sglarge achieves the best average score among the compared models at 88.2, improving over \qwensglargestrict by 1.5 F1. StreamGuard is especially strong on \aegisii and \wgtest, while Qwen3Guard loose variants remain better on \toxicchat and \oaimod, likely reflecting policy mismatch. Overall, StreamGuard remains a strong input moderator while matching or exceeding prior streaming guardrails on standard input benchmarks.

\paragraph{Output moderation.}
Table~\ref{table2} reports response-level F1 under the streaming protocol of Section~\ref{sec:dataset_metrics}. \sglarge achieves the best average among streaming models at 81.9 F1, outperforming \qwensglargestrict by 1.5 points. It is strongest on \srlhf, \xstestresp, and \aegisii, while \qwensg remains better on \harmbench. Overall, forecasting future risk yields a strong streaming output moderator and improves over prior streaming baselines while remaining competitive with offline judges that operate on complete responses.

\paragraph{Streaming intervention and overblocking.}
Table~\ref{tab:detection-latency} reports intervention timing on \qwenstreamtest \responseloc and overblocking on \xstestresp. On \responseloc, all examples are unsafe, so precision is always 100 and F1 differences reflect recall. Across model sizes, \streamguard consistently improves F1 and OnTime while reducing Miss relative to the corresponding \qwensg baselines. At 8B, \sglarge reaches 97.5 F1 and 92.6\% OnTime, compared to 95.9 F1 and 89.9\% for \qwensglargestrict, while reducing Miss from 7.9\% to 4.9\%.

These gains do not come from indiscriminate triggering. On \xstestresp, \sglarge gives the best 8B trade-off at 89.9 F1 with 4.5 FPR, slightly improving on \qwensglargestrict. Overall, forecasting improves timely intervention while preserving benign calibration on challenging safe responses.

\begin{table*}[]
\centering
\small
\resizebox{\textwidth}{!}{%
\begin{tabular}{@{}lllllll@{}}
\toprule
\multirow{2}{*}{Model}               & \multicolumn{4}{c}{\qwenstreamtest}                                                                                        & \multicolumn{2}{c}{\xstestr}                \\ \cmidrule(rl){2-5}  \cmidrule(rl){6-7} 
                                     & F1                              & OnTime (\%)                     & Late (\%)             & Miss (\%)                      & F1                     & FPR                   \\ \midrule
\qwensgsmallstrict                  & 95.0          & 88.3          & 2.1 & 9.6          & 83.8                   & 7.1                   \\
\qwensgsmallloose                   & 94.3          & 87.9          & 1.2 & 10.8         & 83.8                   & 7.1                   \\
\sgsmall (\textbf{ours})            & \textbf{97.2\small{$\pm\,0.2$}} & \textbf{92.9\small{$\pm\,0.5$}} & 1.6\small{$\pm\,0.2$} & \textbf{5.5\small{$\pm\,0.4$}} & \textbf{84.6\small{$\pm{0.7}$}} & \textbf{6.3\small{$\pm{0.4}$}} \\ \cmidrule(rl){1-7} 
\qwensgmediumstrict                 & 96.8         & 91.0         & 2.7         & 6.3         & 87.7                   & \textbf{4.9}                   \\
\qwensgmediumloose                  & 94.9         & 88.3         & 2.0         & 9.7          & \textbf{88.0}                     & 5.4                   \\
\sgmedium (\textbf{ours})           & \textbf{97.6\small{$\pm\,0.0$}} & \textbf{92.4\small{$\pm\,0.3$}} & 2.8\small{$\pm\,0.2$} & \textbf{4.8\small{$\pm\,0.1$}} & 87.2\small{$\pm{0.2}$} & 6.0\small{$\pm{0.2}$} \\  \cmidrule(rl){1-7} 
\qwensglargestrict                  & 95.9          & 89.9          & 2.2 & 7.9          & 88.9                   & 4.6                   \\
\qwensglargeloose                   & 95.0          & 88.7          & 1.7 & 9.6          & 88.0                     & 5.4                   \\
\sglarge (\textbf{ours})            & \textbf{97.5\small{$\pm\,0.0$}} & \textbf{92.6\small{$\pm\,0.2$}} & 2.5\small{$\pm\,0.2$} & \textbf{4.9\small{$\pm\,0.0$}} & \textbf{89.9\small{$\pm{0.5}$}} & \textbf{4.5\small{$\pm{0.3}$}} \\ \bottomrule
\end{tabular}%
}
\caption{Streaming safety evaluation on \qwenstreamtest and overblocking on \xstestresp. Interventions are categorized as \texttt{OnTime} (at or before the unsafe sentence ends), \texttt{Late} (after the sentence ends), or \texttt{Miss} (never triggered) (\S~\ref{sec:dataset_metrics}). \xstestresp reports response F1 and benign FPR. \streamguard models report mean $\pm$ std over three seeds.}
\label{tab:detection-latency}
\end{table*}

\begin{table*}[t]
\centering
\resizebox{\textwidth}{!}{%
\begin{tabular}{@{}llccllllll@{}}
\toprule
\multirow{2}{*}{Model}        & \multirow{2}{*}{X-Tok} & Input Mod    & Output Mod  & \multicolumn{4}{c}{\qwenstreamtest}                                                                                         & \multicolumn{2}{c}{\xstestr}                                  \\ \cmidrule(rl){3-3} \cmidrule(rl){4-4} \cmidrule(rl){5-8}  \cmidrule(rl){9-10} 
                              &                                  & F1 Mean       & F1 Mean       & F1                              & OnTime (\%)                     & Late (\%)              & Miss (\%)                      & F1                              & FPR                            \\ \midrule
\qwensgsmallstrict           & No                               & 84.7          & 78.8          & 95.0                            & 88.3                            & 2.1                    & 9.6                            & 83.8                            & 7.1                            \\
\qwensgsmallloose            & No                               & 84.4          & 77.8          & 94.3                            & 87.9                            & 1.2                    & 10.8                           & 83.8                            & 7.1                            \\
\sgsmall                     & No                               & \textbf{86.4} & 80.2          & 97.2 \small{$\pm  0.2$}         & 92.9 \small{$\pm  0.5$}         & 1.6 \small{$\pm  0.2$} & 5.5 \small{$\pm  0.4 $}        & 84.6 \small{$\pm 0.7$}            & 6.3 \small{$\pm 0.4$}          \\ \cmidrule(rl){1-10} 
\sggemmasmall                & Yes                              & 79.9          & 75.3          & 97.6 \small{$\pm 0.4$}          & 91.3 \small{$\pm 0.4$}          & 4.0 \small{$\pm 0.4$}  & 4.8 \small{$\pm 0.8$}          & 76.1 \small{$\pm 2.4$}          & 11.6 \small{$\pm 1.2$}         \\
\sggemmamedium               & Yes                              & 85.1          & \textbf{81.3} & \textbf{98.2 \small{$\pm 0.2$}} & \textbf{92.3 \small{$\pm 0.3$}} & 4.2 \small{$\pm 0.1$}  & \textbf{3.5 \small{$\pm 0.4$}} & \textbf{87.2 \small{$\pm 0.0$}} & \textbf{5.2 \small{$\pm 0.0$}} \\
\sgqwensmall                 & Yes                              & 82.3          & 79.4          & 97.6 \small{$\pm 0.3$}          & 92.3 \small{$\pm 0.4$}          & 3.0 \small{$\pm 0.1$}  & 4.7 \small{$\pm 0.6$}          & 82.8 \small{$\pm 0.8$}          & 7.2 \small{$\pm 0.4$}          \\
\sgqwenmedium                & Yes                              & 85.8          & 80.1          & 98.0 \small{$\pm 0.2$}          & 92.3 \small{$\pm 0.3$}          & 3.8 \small{$\pm 0.2$}  & 3.9 \small{$\pm 0.3$}          & 83.0 \small{$\pm 1.7$}          & 8.2 \small{$\pm 0.8$}          \\ \bottomrule
\end{tabular}%
}
\caption{Transfer across tokenizers and backbone families. \texttt{X-Tok} denotes prefix-level supervision transferred from a source with a different tokenizer. Input and Output Mod report macro-averaged F1 across standard suites (dataset-level breakdowns in Appendix~\ref{app:cross-tokenizer}). \qwenstreamtest and \xstestresp protocols follow \S~\ref{sec:dataset_metrics}. \streamguard models report mean $\pm$ std over three seeds.}
\label{tab:cross-tokenizer-results}
\end{table*}

\paragraph{Cross-tokenizer transfer.}
Table~\ref{tab:cross-tokenizer-results} shows that forecasting-based supervision transfers effectively across tokenizers and model families. On the output side, the strongest transferred model is \sggemmamedium, which reaches 81.3 output-moderation F1, 98.2 streaming F1, and a 3.5\% miss rate. Relative to the native \sgsmall baseline, it improves output-moderation average, streaming F1, miss rate, and \xstestr F1, though its OnTime rate is slightly lower.
On the input side, \sgqwenmedium is strongest among the transferred models at 85.8. Input moderation varies more across families, but the overall pattern is clear: future-risk supervision remains effective even when transferred across tokenizer families.

\paragraph{Wall-clock latency.}
Appendix~\ref{app:latency}, Table~\ref{tab:latency}, reports wall-clock latency for streaming moderation. Across the evaluated guardrails, decision latency ranges from 2.4\,ms to 9.5\,ms, corresponding to 105--420 decisions/s. Relative to Llama3-8B-Instruct, decision ratios range from 0.243 to 0.969, and relative to Llama3-70B-Instruct they range from 0.024 to 0.096. Operationally, this means that once a block decision is produced, the measured pairings stay below the regime where another token would typically be exposed to the user.

\section{Ablation Studies}
\label{sec:results_ablations}

Table~\ref{tab:ablations} studies three practical target-construction choices with the \sglarge backbone: reduction rule, rollout-source composition, and supervision density. Overall, \streamguard is robust to these choices. Input moderation changes little across variants, while the main trade-off is between early intervention and better benign calibration.

\paragraph{Reduction.}
Reduction has the clearest effect on operating point. Max reduction gives the strongest streaming numbers (98.1 F1, 95.1\% OnTime, 3.8\% Miss) but hurts overall output moderation and benign calibration, whereas min reduction is more conservative and lowers FPR at the cost of substantially worse streaming performance. Mean reduction provides the best overall balance and is therefore used as the reference setup.

\paragraph{Rollout mix.}
Rollout-source composition changes the balance between aggressiveness and robustness. Single-model Llama rollouts slightly improve streaming intervention, but mixed-generator rollouts yield better overall output moderation and calibration. This supports using a diverse rollout pool in the reference configuration.

\paragraph{Supervision density.}
Streaming performance is relatively insensitive to supervision density, suggesting that dense labels at every token are unnecessary. A modest dense prefix with a sparse tail preserves the main gains while improving efficiency, and D64 S4 gives the best \xstestresp trade-off in this block.

\begin{table*}[t]
\centering
\resizebox{0.95\textwidth}{!}{%
\begin{tabular}{@{}llccllllll@{}}
\toprule
\multirow{2}{*}{Factor}               & \multirow{2}{*}{Variant}  & \multicolumn{1}{l}{Input Mod} & \multicolumn{1}{l}{Output Mod} & \multicolumn{4}{c}{\qwenstreamtest}                                                             & \multicolumn{2}{c}{\xstestr}                \\ \cmidrule(rl){3-3}  \cmidrule(rl){4-4} \cmidrule(rl){5-8}  \cmidrule(rl){9-10} 
                                      &                           & F1 Mean                      & F1 Mean                    & F1                     & OnTime (\%)            & Late (\%)             & Miss (\%)             & F1                     & FPR                   \\ \midrule
Reference                             & Mean Reduction            & 88.2                           & 81.9                         & 97.5\small{$\pm\,0.0$} & 92.6\small{$\pm\,0.2$} & 2.5\small{$\pm\,0.2$} & 4.9\small{$\pm\,0.0$} & 89.9\small{$\pm\,0.5$} & 4.5\small{$\pm\,0.3$}                     \\ \cmidrule(rl){1-10} 
\multirow{4}{*}{Reduction}            & Beta Mean                 & 88.2                           & 81.4                         & 96.7 \small{$\pm 0.1$} & 90.3 \small{$\pm 1.0$} & 3.3 \small{$\pm 1.0$} & 6.4 \small{$\pm 0.1$} & 89.2 \small{$\pm 0.7$} & 3.9 \small{$\pm 0.6$} \\
                                      & Median                    & 88.1                           & 81.2                         & 96.8 \small{$\pm 0.5$} & 90.1 \small{$\pm 1.5$} & 3.6 \small{$\pm 0.6$} & 6.3 \small{$\pm 0.9$} & 88.6 \small{$\pm 0.2$} & 4.8 \small{$\pm 0.6$} \\
                                      & Max                       & 88.1                           & 79.0                         & 98.1 \small{$\pm 0.0$} & 95.1 \small{$\pm 0.2$} & 1.2 \small{$\pm 0.1$} & 3.8 \small{$\pm 0.1$} & 85.9 \small{$\pm 0.2$} & 6.8 \small{$\pm 0.4$} \\
                                      & Min                       & 87.7                           & 80.9                         & 95.5 \small{$\pm 0.3$} & 82.5 \small{$\pm 0.7$} & 9.0 \small{$\pm 1.3$} & 8.6 \small{$\pm 0.6$} & 88.4 \small{$\pm 0.9$} & 4.1 \small{$\pm 0.3$} \\ \midrule
\multirow{3}{*}{Rollout Mix}     & Two-model mixture (32)    & 88.1                           & 81.1                         & 96.6 \small{$\pm 0.1$} & 90.3 \small{$\pm 0.6$} & 3.1 \small{$\pm 0.4$} & 6.5 \small{$\pm 0.2$} & 89.4 \small{$\pm 0.0$} & 4.3 \small{$\pm 0.0$} \\
                                      & Single-model (llama3-8B)  & 88.2                           & 80.9                         & 97.6 \small{$\pm 0.1$} & 93.8 \small{$\pm 0.3$} & 1.6 \small{$\pm 0.0$} & 4.6 \small{$\pm 0.3$} & 88.5 \small{$\pm 2.3$} & 5.2 \small{$\pm 0.8$} \\
                                      & Single-model (qwen2.5 7B) & 88.0                           & 80.4                         & 95.9 \small{$\pm 0.1$} & 85.9 \small{$\pm 1.6$} & 6.3 \small{$\pm 1.6$} & 7.8 \small{$\pm 0.2$} & 88.6 \small{$\pm 2.0$} & 4.4 \small{$\pm 0.7$} \\ \midrule
\multirow{8}{*}{Sup Density} & D64 S4                    & 88.3                           & 82.0                         & 97.5                   & 92.7                   & 2.3                   & 4.9                   & 90.6                   & 4.1                   \\
                                      & D64 S8                    & 88.5                           & 81.9                         & 97.6                   & 92.9                   & 2.5                   & 4.7                   & 89.7                   & 4.9                   \\
                                      & D32 S4                    & 88.3                           & 81.6                         & 97.8                   & 93.4                   & 2.3                   & 4.3                   & 89.1                   & 5.2                   \\
                                      & D32 S8                    & 88.5                           & 82.0                         & 97.5                   & 92.4                   & 2.7                   & 4.9                   & 90.2                   & 4.6                   \\
                                      & D16 S4                    & 88.3                           & 81.9                         & 97.7                   & 92.6                   & 2.8                   & 4.6                   & 89.5                   & 4.6                   \\
                                      & D16 S8                    & 88.4                           & 82.0                         & 97.7                   & 92.5                   & 3.1                   & 4.4                   & 89.5                   & 4.6                   \\
                                      & D1 S4                     & 88.1                           & 81.1                         & 97.7                   & 93.5                   & 2.1                   & 4.4                   & 88.1                   & 5.7                   \\
                                      & D1 S8                     & 88.4                           & 81.9                         & 97.7                   & 93.1                   & 2.3                   & 4.6                   & 89.5                   & 4.6                   \\ \bottomrule
\end{tabular}%
}
\caption{Unified ablation results for \streamguard with the Llama-3-8B backbone. \textbf{Reduction}, \textbf{Rollout Mix}, and \textbf{Sup Density} vary the rollout-score aggregation rule, rollout-generator composition, and dense-prefix / strided-tail supervision schedule, respectively; \textbf{Reference} is the main configuration used in the paper. Prompt Mod and Resp Mod report macro-averaged F1 over the standard benchmark suites; Appendix~\ref{app:ablations} provides the dataset-level moderation breakdowns. \qwenstreamtest and \xstestresp use the same streaming and overblocking protocols as in \S~\ref{sec:dataset_metrics}. We report mean $\pm$ std over three seeds where shown.}
\label{tab:ablations}
\end{table*}

\section{Related Work}
\label{sec:related-work}

\paragraph{Full-Sequence Output Moderation.}
Most deployed guardrails operate \emph{post hoc}: they inspect a complete prompt or response after it has already been produced. Models such as LlamaGuard~\citep{inan2023llamaguardllmbasedinputoutput} and WildGuard~\citep{NEURIPS2024_wildguard} cast safety as full-sequence classification over a fixed taxonomy of harms. These systems are effective as offline judges, but they do not natively solve the streaming problem because they require the completed response before producing a decision. Recent moderators such as BingoGuard~\citep{bingoguard2025} and policy-adaptive systems such as DynaGuard~\citep{dynaguard2025} and OpenAI's gpt-oss-safeguard~\citep{openai2025safeguard} increase granularity or flexibility, but they still rely on full context and therefore inherit the same latency bottleneck in streaming settings.

\paragraph{White-Box and Internal Defenses.}
A separate line of work reduces moderation latency by leveraging the generator LLM's internal activations. ShieldHead~\citep{shieldhead} and related internal probes attach lightweight safety heads to hidden states, while recent streaming white-box systems such as Kelp~\citep{plugguard} extend this idea to token-level intervention during decoding by reading internal activations online and modeling risk over time. These methods can be fast, but they are architecture-dependent: the detector is trained on the representation space of a particular base model and is therefore coupled to that model family and checkpoint. In contrast, StreamGuard is a text-only guardrail trained and deployed over prefixes in text space, which makes it compatible with standard black-box moderation setups.

\paragraph{Streaming Safety Systems.}
Streaming-specific safety systems are the closest prior work to our setting because they moderate partial output prefixes online. Constitutional Classifiers~\citep{constitutionalclassifiers2025} are conceptually close, as they also train on partial outputs to anticipate eventual harmfulness. However, they use prefix-to-final-label supervision, whereas StreamGuard supervises each prefix with a Monte Carlo estimate of the expected harmfulness of likely future continuations. Because the system is not publicly released, we treat it as related work rather than as a direct empirical baseline. In the open text-based setting, \qwensg~\citep{qwen3guard2025} is the strongest directly comparable baseline, with publicly released models; however, it uses a different supervision signal, casting streaming moderation as boundary detection by identifying the earliest token at which the response becomes unsafe, labeling earlier prefixes as safe and the boundary and all later prefixes as unsafe or controversial. StreamGuard differs from both approaches in its training target: rather than assigning each prefix the label of one realized completion or the label induced by an explicit unsafe boundary, it supervises each observed prefix with a Monte Carlo estimate of the expected harmfulness of likely future continuations. As a result, StreamGuard is trained to forecast future risk from the current prefix, not merely to inherit a final-output label or detect that an unsafe boundary has already been crossed.

\section{Conclusion}

We introduced \streamguard, a unified text-based guardrail for input moderation and streaming output moderation. For streaming outputs, \streamguard treats moderation as a forecasting problem: given a partial prefix, it predicts the harmfulness of likely future continuations rather than waiting until an explicit violation has already appeared. This enables earlier intervention during generation without requiring exact boundary annotations.

Across standard moderation benchmarks, \streamguard remains competitive as a conventional moderator while improving over strong published streaming baselines on output-side and intervention-timing evaluations. On the streaming benchmark, it achieves higher overall recall, more on-time interventions, and fewer missed unsafe responses than the comparable boundary-based baseline. We also find that forecasting-based supervision transfers well across tokenizer families, yielding strong output-moderation and streaming performance on Gemma and Qwen backbones.
Overall, our results suggest that forecasting is a useful and practical framing for streaming safety: it is simple to train, compatible with text-only black-box moderation, and also effective in low-latency deployment settings.

\section*{Acknowledgments}
The authors gratefully acknowledge Shuta Hoshino for their contribution to the early development of this project.

\bibliography{reference}
\bibliographystyle{colm2026_conference}

\clearpage
\appendix

\section{Training and Transfer Details}
\label{app:training-transfer-details}

This appendix provides the full training and transfer setup underlying the main results. We repeat the core setup from the main text for completeness and add the optimization and implementation details omitted there for space.

\subsection{Backbones and Transfer Setup}

We instantiate \streamguard across multiple backbone families and sizes. Our primary scaling experiments use Llama3 backbones at 1B, 3B, and 8B. To study portability beyond a single tokenizer family, we additionally evaluate transfer to Gemma and Qwen backbones at small model scales.

Because our prefix-level supervision is primarily constructed with the Llama3 tokenizer, we study transfer by applying these Llama3-based targets to Gemma and Qwen backbones. In the native setting, prefix targets can be consumed directly as tokenizer-aligned token labels. In the transfer setting, where tokenizer coordinates do not match, the same supervision is reused in text space by pairing decoded prefixes with future-risk targets. This contrasts with boundary-based streaming supervision, where unsafe onset indices are tied to tokenizer-specific coordinates and are therefore not directly portable across architectures. We compare this transferred setting against a native same-tokenizer setting.

\subsection{Reference Training and Target Construction}

Training and target construction use examples from \wgtrain, \aegisii, \toxicchat, and \beavertails. Unless otherwise noted, all main results use a fixed reference StreamGuard target-construction setup. This includes mean reduction, a four-model rollout pool, and a fixed supervision schedule.

For the reference rollout pool, we sample continuations from Llama~3~\citep{grattafiori2024llama3herdmodels} 8B and 70B, Qwen2.5~\citep{qwen2025qwen25technicalreport} 7B and 72B, drawing four rollouts per model at temperature 0.7. We score the rollouts using \qwenguardstrict~\citep{qwen3guard2025}. Rollout expansion is only required for unsafe prompts, since unsafe responses arise only when the corresponding prompts are unsafe.

Because rollout-based target construction is expensive for long responses, our reference training setup uses a budgeted supervision schedule on longer-response datasets such as \wgtrain, applying prefix supervision densely over an initial segment and more sparsely thereafter. In the main paper, rollout-source composition, reduction, and supervision density are treated as practical implementation choices and analyzed in Section~\ref{sec:results_ablations}.

\subsection{Implementation Details}

\paragraph{Training Configuration.}
We train all models with a maximum sequence length of 8192 tokens and an effective global batch size of 128, using gradient accumulation as needed. We determine the learning rate through a grid search over $\{2e\text{-}6, 5e\text{-}6, 8e\text{-}6, 1e\text{-}5\}$, selecting $8e\text{-}6$ for 8B models and $1e\text{-}5$ for smaller models. Optimization uses AdamW with default hyperparameters except for the learning rate, together with a \texttt{StepLR} scheduler with step size 1 epoch and decay factor $\gamma = 0.85$. All models are trained in bfloat16 with full-parameter fine-tuning for up to three epochs, using gradient clipping with a maximum norm of 1.0 and no warmup. We merge all training datasets into a single training corpus. Checkpoints are selected based on F1 on the \aegisii official validation set~\citep{ghosh2025aegis20diverseaisafety}. All experiments use custom PyTorch~\citep{paszke2019pytorchimperativestylehighperformance} training code with FSDP~\citep{zhao2023pytorchfsdpexperiencesscaling}. Unless otherwise noted, each model variant is trained with three random seeds, and we report the mean and standard deviation across runs.

\paragraph{Architecture and Inference.}
We attach a linear classification head directly to the final hidden state of every token in the sequence. This enables the model to produce a continuous risk estimate $V_\theta(x, y_{\le t})$ at every step of the generation without altering the input structure. At test time, a prefix is classified as unsafe when its predicted score exceeds a threshold, which defaults to 0.5, with the final trigger determined by the debouncing rule described in Section~\ref{sec:dataset_metrics}.

\section{Full Standard Moderation Results}
\label{app:full-standard-results}

For space, the main text reports all streaming baselines together with a reduced set of offline references. This appendix restores the full baseline panel for the standard input- and output-moderation comparisons. These expanded tables preserve the same overall conclusions as the main text: \streamguard remains competitive with strong offline post-hoc moderators while improving over prior streaming baselines in the setting most relevant to this paper.

\subsection{Input Moderation}
\label{app:full-input-moderation}

Table~\ref{table1-full} reports the full input-moderation comparison, including all offline and streaming baselines. The main text keeps the discussion focused on the strongest and most relevant comparisons for the streaming setting, while this appendix restores the complete model set for transparency. The full table confirms that the main-text pattern is not an artifact of baseline selection: \streamguard remains strong on the macro-average and particularly competitive on newer safety benchmarks such as \aegisii and \wgtest, while the looser Qwen variants remain relatively stronger on \toxicchat and \oaimod, consistent with the policy-mismatch.

\begin{table*}[]
\centering
\resizebox{\textwidth}{!}{%
\begin{tabular}{@{}llllllllll@{}}
\toprule
Model                                 & Streaming & ToxiC                  & OAIMod                 & Aegis                           & Aegis2                          & SSTest                          & HarmB                  & WildG                           & Avg           \\ \midrule
LlamaGuard3-8B                        & no        & 53.8                   & 79.5                   & 71.5                            & 76.4                            & 99.5                            & 99.0                   & 76.4                            & 79.4          \\
LlamaGuard4-12B                       & no        & 51.3                   & 73.5                   & 67.8                            & 70.6                            & 98.0                            & 97.2                   & 73.0                            & 75.9          \\
WildGuard-7B                          & no        & 70.8                   & 72.1                   & 89.4                            & 80.7                            & 99.5                            & 98.9                   & 88.9                            & 85.8          \\
ShieldGemma-9B~\citep{zeng2024shieldgemmagenerativeaicontent}                        & no        & 69.4                   & 82.1                   & 70.3                            & 72.5                            & 83.7                            & 60.6                   & 54.2                            & 70.4          \\
ShieldGemma-27B                       & no        & 72.9                   & 80.5                   & 69.0                            & 71.6                            & 84.4                            & 57.3                   & 54.3                            & 70.0          \\
NemoGuard-8B                          & no        & 75.6                   & 81.0                   & 81.4                            & 86.8                            & 98.5                            & 75.2                   & 81.6                            & 82.9          \\
PolyGuard-Qwen-7B~\citep{kumar2025polyguardmultilingualsafetymoderation}                     & no        & 71.5                   & 74.1                   & 90.3                            & 86.3                            & 100.0                           & 98.7                   & 88.1                            & 87.0          \\
Qwen3Guard-0.6B-Gen-strict            & no        & 65.1                   & 66.5                   & 90.8                            & 85.0                            & 99.0                            & 98.7                   & 87.7                            & 84.7          \\
Qwen3Guard-0.6B-Gen-loose             & no        & 77.7                   & 77.6                   & 76.9                            & 83.3                            & 95.8                            & 96.1                   & 85.1                            & 84.6          \\
Qwen3Guard-4B-Gen-strict              & no        & 69.5                   & 68.3                   & 90.8                            & 85.8                            & 99.5                            & 100.0                  & 85.6                            & 85.6          \\
Qwen3Guard-4B-Gen-loose               & no        & 82.8                   & 80.7                   & 76.3                            & 82.1                            & 97.4                            & 99.2                   & 85.1                            & 86.2          \\
Qwen3Guard-8B-Gen-strict              & no        & 68.9                   & 68.8                   & 91.4                            & 86.1                            & 99.5                            & 100.0                  & 88.9                            & 86.2          \\
Qwen3Guard-8B-Gen-loose               & no        & 82.8                   & 81.3                   & 76.0                            & 82.5                            & 97.4                            & 98.5                   & 85.6                            & 86.3          \\ \cmidrule(rl){1-10}
\qwensgsmallstrict                    & yes       & 72.0                   & 68.3                   & 85.2                            & 84.9                            & 98.0                            & 97.2                   & 87.1                            & 84.7          \\
\qwensgsmallloose                     & yes       & 75.5                   & 76.0                   & 77.7                            & 81.7                            & 96.9                            & 96.8                   & 86.0                            & 84.4          \\
\qwensgmediumstrict                   & yes       & 73.0                   & 70.0                   & 85.9                            & 86.6                            & \textbf{99.5}                   & \textbf{100.0}         & 88.6                            & 86.2          \\
\qwensgmediumloose                    & yes       & 81.7                   & 81.2                   & 75.5                            & 80.2                            & 98.5                            & 98.9                   & 85.3                            & 85.9          \\
\qwensglargestrict                    & yes       & 75.3                   & 74.0                   & 85.7                            & 86.1                            & 99.0                            & 99.4                   & 87.5                            & 86.7          \\
\qwensglargeloose                     & yes       & \textbf{80.1}          & \textbf{80.3}          & 75.5                            & 80.8                            & 98.5                            & 98.7                   & 84.4                            & 85.5          \\
\sgsmall \textbf{(ours)}              & yes       & 74.9\small{$\pm\,0.3$} & 71.5\small{$\pm\,0.2$} & \textbf{89.0\small{$\pm\,0.3$}} & 87.7\small{$\pm\,0.2$}          & 98.6\small{$\pm\,0.2$}          & 94.8\small{$\pm\,2.9$} & 88.6\small{$\pm\,0.2$}          & 86.4          \\
\sgmedium \textbf{(ours)}             & yes       & 77.7\small{$\pm\,0.5$} & 74.4\small{$\pm\,0.3$} & 87.1\small{$\pm\,0.4$}          & 87.8\small{$\pm\,0.1$}          & 99.3\small{$\pm\,0.2$}          & 99.2\small{$\pm\,0.5$} & 89.0\small{$\pm\,0.2$}          & 87.8          \\
\sglarge \textbf{(ours)}              & yes       & 77.4\small{$\pm\,0.5$} & 75.0\small{$\pm\,0.4$} & 88.5\small{$\pm\,0.2$}    & \textbf{87.9\small{$\pm\,0.1$}} & \textbf{99.5\small{$\pm\,0.0$}} & 99.7\small{$\pm\,0.1$} & \textbf{89.5\small{$\pm\,0.2$}} & \textbf{88.2} \\ \bottomrule
\end{tabular}%
}
\caption{Input-moderation F1; Avg denotes the macro-average. For prior systems, we report published results. We report mean $\pm$ std over three seeds for \streamguard models. The \textit{Streaming} column indicates whether a model supports token-level streaming moderation.}
\label{table1-full}
\end{table*}

\subsection{Output Moderation}
\label{app:full-output-moderation}

Table~\ref{table2-full} reports the full output-moderation comparison under the same streaming response-level protocol used in the main text. As with input moderation, the main paper emphasizes the comparisons most central to our claim---namely, how \streamguard compares to prior streaming guardrails and to strong offline post-hoc references---while this appendix restores the full baseline set. The expanded table confirms the same qualitative conclusion as the main text: \streamguard remains among the strongest models in the causal streaming setting, and its gains over prior streaming baselines are not explained by selective baseline reporting.

\begin{table*}[]
\centering
\resizebox{\textwidth}{!}{%
\begin{tabular}{@{}lllllllll@{}}
\toprule
Model                                 & Streaming & HarmB                  & SRLHF                           & BeaverTails                     & \xstestr                         & Aegis2                          & WildG                           & Avg           \\ \midrule
LlamaGuard3-8B                        & no        & 84.5                   & 45.2                            & 67.9                            & 89.8                            & 66.1                            & 69.5                            & 70.5          \\
LlamaGuard4-12B                       & no        & 83.3                   & 42.5                            & 68.6                            & 88.9                            & 63.7                            & 66.4                            & 68.9          \\
WildGuard-7B                          & no        & 86.3                   & 64.2                            & 84.4                            & 94.7                            & 83.2                            & 75.4                            & 81.4          \\
ShieldGemma-9B                        & no        & 60.4                   & 44.2                            & 62.4                            & 86.3                            & 70.8                            & 49.9                            & 62.3          \\
ShieldGemma-27B                       & no        & 62.9                   & 52.6                            & 67.6                            & 83.0                            & 74.9                            & 52.4                            & 65.6          \\
NemoGuard-8B                          & no        & 81.4                   & 57.6                            & 78.5                            & 86.2                            & 87.6                            & 77.5                            & 78.1          \\
PolyGuard-Qwen-7B                     & no        & 71.1                   & 63.3                            & 79.5                            & 63.4                            & 81.9                            & 77.9                            & 72.9          \\
Qwen3Guard-0.6B-Gen-strict            & no        & 85.0                   & 66.6                            & 86.1                            & 89.7                            & 84.2                            & 76.3                            & 81.3          \\
Qwen3Guard-0.6B-Gen-loose             & no        & 82.6                   & 64.2                            & 85.4                            & 91.3                            & 84.1                            & 77.3                            & 80.8          \\
Qwen3Guard-4B-Gen-strict              & no        & 86.7                   & 69.8                            & 86.6                            & 92.7                            & 86.1                            & 79.5                            & 83.6          \\
Qwen3Guard-4B-Gen-loose               & no        & 86.7                   & 64.5                            & 85.2                            & 92.4                            & 86.5                            & 77.3                            & 82.1          \\
Qwen3Guard-8B-Gen-strict              & no        & 87.2                   & 70.5                            & 86.6                            & 92.1                            & 86.1                            & 78.9                            & 83.6          \\
Qwen3Guard-8B-Gen-loose               & no        & 86.5                   & 64.2                            & 85.5                            & 93.7                            & 86.4                            & 77.3                            & 82.3          \\ \cmidrule(rl){1-9}
\qwensgsmallstrict                    & yes       & 83.1                   & 62.8                            & 84.5                            & 84.8                            & 81.4                            & 76.3                            & 78.8          \\
\qwensgsmallloose                     & yes       & 80.6                   & 61.7                            & 84.0                            & 83.3                            & 81.4                            & 75.8                            & 77.8          \\
\qwensgmediumstrict                   & yes       & 84.3                   & 67.6                            & 86.0                            & 88.5                            & 83.1                            & 76.4                            & 81.0          \\
\qwensgmediumloose                    & yes       & 83.6                   & 64.3                            & 85.2                            & 88.9                            & 83.3                            & 77.4                            & 80.5          \\
\qwensglargestrict                    & yes       & \textbf{85.0}          & 64.6                            & 85.9                            & 87.5                            & 82.6                            & 77.0                            & 80.4          \\
\qwensglargeloose                     & yes       & 84.7                   & 63.1                            & 85.5                            & 88.9                            & 82.4                            & 76.8                            & 80.2          \\
\sgsmall \textbf{(ours)}              & yes       & 82.4\small{$\pm\,0.2$} & 68.0\small{$\pm\,0.7$}          & 86.2\small{$\pm\,0.1$}          & 84.6\small{$\pm\,0.7$}          & 82.9\small{$\pm\,0.3$}          & 77.2\small{$\pm\,0.3$}          & 80.2          \\
\sgmedium \textbf{(ours)}             & yes       & 83.2\small{$\pm\,0.3$} & 68.2\small{$\pm\,0.2$}          & 86.4\small{$\pm\,0.0$}          & 87.2\small{$\pm\,0.2$}          & 83.2\small{$\pm\,0.2$}          & \textbf{77.8\small{$\pm\,0.3$}} & 81.0          \\
\sglarge \textbf{(ours)}              & yes       & 83.3\small{$\pm\,0.3$} & \textbf{69.2\small{$\pm\,0.3$}} & \textbf{86.6\small{$\pm\,0.2$}} & \textbf{89.9\small{$\pm\,0.5$}} & \textbf{84.5\small{$\pm\,0.1$}} & \textbf{77.8\small{$\pm\,0.1$}} & \textbf{81.9} \\ \bottomrule
\end{tabular}%
}
\caption{Output-moderation F1; Avg denotes the macro-average. For prior systems, we report published results. \streamguard models are evaluated in our simulated streaming protocol, processing responses one token at a time and counting a response as unsafe if it triggers under the debounced rule in \S~\ref{sec:dataset_metrics}. We report mean $\pm$ std over three seeds for \streamguard models. The \textit{Streaming} column indicates whether a model supports token-level moderation during generation.}
\label{table2-full}
\end{table*}

\section{Dataset-Level Ablation Details}
\label{app:ablations}

This appendix unpacks the aggregate ablation results in Table~\ref{tab:ablations} into per-benchmark moderation scores. We retain the same \sglarge backbone and the same three practical target-construction factors studied in Section~\ref{sec:results_ablations}: the reduction rule used to aggregate rollout safety scores, the rollout-generator mix across the compared single-model, two-model, and four-model settings, and the supervision-density schedule. These appendix tables should be read as complements to the unified main-text ablation table rather than as separate experiments: Table~\ref{tab:ablations} remains the summary view for streaming intervention and overblocking, while the appendix tables show which standard moderation benchmarks drive the small average differences across variants. In particular, the input-side table is most useful for checking that no single benchmark dominates the macro-average, whereas the output-side table makes clear which datasets absorb the cost of more aggressive streaming operating points.

\subsection{Input Moderation}
Table~\ref{tab:ablations-input} reports dataset-level input-moderation F1 for the ablation variants. The central pattern is stability: across reduction rules and rollout mixtures, the macro-average stays in a narrow 87.7--88.2 band, which shows that the small input-side differences in Table~\ref{tab:ablations} are not driven by a single benchmark. No reduction rule dominates the full suite.

Within the reduction block, median reduction improves Aegis to 89.9 and attains the strongest WildGuard score among the reduction variants at 89.6, but gives up performance on \oaimod; min reduction is the weakest overall reduction variant, largely because it drops on \toxicchat (75.9) and \oaimod (73.0). Rollout-mix changes mostly reshuffle where the gains appear: single-model \texttt{llama3-8B} rollouts slightly improve Aegis (89.5) and WildGuard (90.0), while the four-model reference remains stronger on \toxicchat and \oaimod. The supervision-density block produces the largest local swings, especially on policy-sensitive datasets such as \toxicchat and \oaimod: D32 S8 reaches 79.0 on \toxicchat, D16 S8 reaches 76.5 on \oaimod, and D1 S8 reaches 90.1 on WildGuard, yet the overall average still remains between 88.1 and 88.5. These details reinforce the main-text interpretation that forecasting-based supervision is robust on standard input moderation, with most practical variants shifting where the model is slightly more or less strict rather than changing the overall quality level.
\begin{table}[]
\centering
\resizebox{\textwidth}{!}{%
\begin{tabular}{@{}llllllllll@{}}
\toprule
Factor               & Variant                   & ToxiC                           & OAIMod                          & Aegis                           & Aegis2                          & SSTest                 & HarmB                  & WildG                           & Avg           \\ \midrule
Reduction            & Mean                      & \textbf{77.4 \small{$\pm\, 0.5$}} & \textbf{75.0 \small{$\pm\, 0.4$}} & 88.5 \small{$\pm\, 0.2$}          & \textbf{87.9 \small{$\pm\, 0.1$}} & 99.5 \small{$\pm\, 0.0$} & 99.7 \small{$\pm\, 0.1$} & 89.5 \small{$\pm\, 0.2$}          & \textbf{88.2} \\
Reduction            & Beta Mean                 & 77.0 \small{$\pm\, 0.9$}          & \textbf{75.0 \small{$\pm\, 2.5$}} & 88.4 \small{$\pm\, 0.3$}          & 87.8 \small{$\pm\, 0.5$}          & 99.5 \small{$\pm\, 0.0$} & 99.9 \small{$\pm\, 0.1$} & 89.5 \small{$\pm\, 0.0$}          & \textbf{88.2} \\
Reduction            & Median                    & 77.0 \small{$\pm\, 2.0$}          & 73.3 \small{$\pm\, 1.3$}          & \textbf{89.9 \small{$\pm\, 0.2$}} & 87.7 \small{$\pm\, 0.2$}          & 99.5 \small{$\pm\, 0.0$} & 99.9 \small{$\pm\, 0.1$} & \textbf{89.6 \small{$\pm\, 0.3$}} & 88.1          \\
Reduction            & Max                       & \textbf{77.4 \small{$\pm\, 0.7$}} & 73.6 \small{$\pm\, 0.2$}          & 89.2 \small{$\pm\, 0.5$}          & \textbf{87.9 \small{$\pm\, 0.5$}} & 99.5 \small{$\pm\, 0.0$} & 99.9 \small{$\pm\, 0.1$} & 89.5 \small{$\pm\, 0.3$}          & 88.1          \\
Reduction            & Min                       & 75.9 \small{$\pm\, 1.7$}          & 73.0 \small{$\pm\, 0.9$}          & 88.4 \small{$\pm\, 0.7$}          & 87.7 \small{$\pm\, 0.1$}          & 99.5 \small{$\pm\, 0.0$} & 99.9 \small{$\pm\, 0.1$} & 89.5 \small{$\pm\, 0.3$}          & 87.7          \\  \cmidrule(rl){1-10}
Rollout Mix          & Four-model mixture (16)   & \textbf{77.4 \small{$\pm\, 0.5$}} & \textbf{75.0 \small{$\pm\, 0.4$}} & 88.5 \small{$\pm\, 0.2$}          & 87.9 \small{$\pm\, 0.1$}          & 99.5 \small{$\pm\, 0.0$} & 99.7 \small{$\pm\, 0.1$} & 89.5 \small{$\pm\, 0.2$}          & \textbf{88.2} \\
Rollout Mix          & Two-model mixture (32)    & 77.1 \small{$\pm\, 0.2$}          & 74.3 \small{$\pm\, 1.9$}          & 88.3 \small{$\pm\, 0.0$}          & \textbf{88.1 \small{$\pm\, 0.1$}} & 99.5 \small{$\pm\, 0.0$} & 99.9 \small{$\pm\, 0.1$} & 89.6 \small{$\pm\, 0.1$}          & 88.1          \\
Rollout Mix          & Single-model (llama3-8B)  & 76.8 \small{$\pm\, 0.3$}          & 73.8 \small{$\pm\, 1.3$}          & \textbf{89.5 \small{$\pm\, 0.7$}} & \textbf{88.1 \small{$\pm\, 0.0$}} & 99.5 \small{$\pm\, 0.0$} & 99.8 \small{$\pm\, 0.0$} & \textbf{90.0 \small{$\pm\, 0.1$}} & \textbf{88.2} \\
Rollout Mix          & Single-model (qwen2.5 7B) & 76.7 \small{$\pm\, 1.5$}          & 73.8 \small{$\pm\, 1.2$}          & 88.6 \small{$\pm\, 1.1$}          & 87.9 \small{$\pm\, 0.1$}          & 99.5 \small{$\pm\, 0.0$} & 99.9 \small{$\pm\, 0.1$} & 89.9 \small{$\pm\, 0.0$}          & 88.0          \\  \cmidrule(rl){1-10}
Supervision Schedule & D64 S\{1,2\}              & 77.4 \small{$\pm\, 0.5$}          & 75.0 \small{$\pm\, 0.4$}          & 88.5 \small{$\pm\, 0.2$}          & 87.9 \small{$\pm\, 0.1$}          & 99.5 \small{$\pm\, 0.0$} & 99.7 \small{$\pm\, 0.1$} & 89.5 \small{$\pm\, 0.2$}          & 88.2          \\
Supervision Schedule & D64 S4                    & 78.0                            & 75.1                            & 88.4                            & 87.8                            & 99.5                   & 99.4                   & 89.9                            & 88.3          \\
Supervision Schedule & D64 S8                    & 78.1                            & 75.9                            & 88.7                            & 87.8                            & 99.5                   & 99.8                   & 89.5                            & \textbf{88.5} \\
Supervision Schedule & D32 S4                    & 77.6                            & 75.1                            & 88.7                            & \textbf{88.0}                   & 99.5                   & \textbf{100.0}         & 89.4                            & 88.3          \\
Supervision Schedule & D32 S8                    & \textbf{79.0}                   & 75.7                            & 88.4                            & 87.9                            & 99.5                   & 99.4                   & 89.7                            & \textbf{88.5} \\
Supervision Schedule & D16 S4                    & 76.7                            & 76.1                            & 88.7                            & 87.7                            & 99.5                   & 100.0                  & 89.7                            & 88.3          \\
Supervision Schedule & D16 S8                    & 79.3                            & \textbf{76.5}                   & 88.1                            & 87.9                            & 99.5                   & 98.3                   & 89.6                            & 88.4          \\
Supervision Schedule & D1 S4                     & 78.4                            & 74.1                            & 88.5                            & 87.8                            & 99.5                   & 99.4                   & 89.2                            & 88.1          \\
Supervision Schedule & D1 S8                     & 76.8                            & 75.3                            & \textbf{89.2}                   & 87.8                            & 99.5                   & \textbf{100.0}         & \textbf{90.1}                   & 88.4          \\ \bottomrule
\end{tabular}%
}
\caption{Dataset-level input-moderation breakdown for the ablations in Table~\ref{tab:ablations}. This appendix table decomposes the aggregate input-moderation result from the main text into per-benchmark F1 scores for the same \sglarge variants, varying the reduction rule used to aggregate rollout oracle scores, the rollout-generator mix across the compared single-model, two-model, and four-model settings, and the supervision-density schedule. Consistent with the paper's main ablation story, input-side moderation remains tightly clustered across these practical target-construction choices, indicating that forecasting-based supervision is robust rather than dependent on a brittle implementation detail. Avg denotes the macro-average across benchmarks; supervision-schedule notation follows Table~\ref{tab:ablations}; reported values are mean $\pm$ standard deviation over three seeds where shown.}
\label{tab:ablations-input}
\end{table}

\subsection{Output Moderation}
Table~\ref{tab:ablations-output} reports dataset-level output-moderation F1 under the same streaming evaluation protocol used throughout the paper. This decomposition makes the main trade-off more concrete.

Within the reduction block, the mean-reduction reference gives the best macro-average (81.9) because it is consistently strong across \beavertails (86.6), \xstest (89.9), and WildGuard (77.8), even though individual alternatives win isolated columns. Max reduction is the clearest example of the aggressive-intervention trade-off: it slightly improves \srlhf to 69.8 and gives the strongest streaming numbers in Table~\ref{tab:ablations}, but gives up substantial performance on \xstest (85.9) and WildGuard (71.2), which explains its weaker 79.0 output-side average. Min reduction shows the opposite operating point, doing best on \harmbench (84.7) and \aegisii (85.0) while dropping sharply on \srlhf (63.3). The rollout-mixture block tells a similar story. The four-model reference is strongest or tied on \srlhf, \beavertails, \xstest, and WildGuard, whereas single-model \texttt{llama3-8B} rollouts slightly improve \harmbench (83.6) but lose on \aegisii and WildGuard; the single-model \texttt{qwen2.5-7B} rollout pool underperforms most clearly on \srlhf (65.7) and WildGuard (75.2).

Supervision density again changes the operating point more than the overall quality: D64 S4 improves \xstest to 90.6 and WildGuard to 78.9, D32 S8 matches the best average at 82.0 with the strongest \beavertails score (86.8), and D1 S4 preserves strong \srlhf performance (69.6) but drops notably on \xstest (88.1) and WildGuard (75.4). Read together with the streaming and FPR columns in Table~\ref{tab:ablations}, these dataset-level shifts make clear that more aggressive variants do not degrade uniformly; they lose most on benchmarks that reward benign calibration or broader output-side robustness.
\begin{table}[]
\centering
\resizebox{\textwidth}{!}{%
\begin{tabular}{@{}lllllllll@{}}
\toprule
Factor               & Variant                   & HarmB                           & SRLHF                           & BeaverTails                     & \xstestr                         & Aegis2                          & WildG                           & Avg           \\ \midrule
Reduction            & Mean                      & 83.3 \small{$\pm\,0.3$}          & 69.2 \small{$\pm\,0.3$}          & \textbf{86.6 \small{$\pm\,0.2$}} & \textbf{89.9 \small{$\pm\,0.5$}} & 84.5 \small{$\pm\,0.1$}          & \textbf{77.8 \small{$\pm\,0.1$}} & \textbf{81.9} \\
Reduction            & Beta Mean                 & 83.4 \small{$\pm 0.7$}          & 67.6 \small{$\pm 0.5$}          & 86.5 \small{$\pm 0.2$}          & 89.2 \small{$\pm 0.7$}          & 84.7 \small{$\pm 0.8$}          & 77.3 \small{$\pm 1.1$}          & 81.4          \\
Reduction            & Median                    & 83.5 \small{$\pm 0.6$}          & 67.3 \small{$\pm 1.0$}          & 86.1 \small{$\pm 0.0$}          & 88.6 \small{$\pm 0.2$}          & 84.3 \small{$\pm 0.9$}          & 77.2 \small{$\pm 0.5$}          & 81.2          \\
Reduction            & Max                       & 79.8 \small{$\pm 0.2$}          & \textbf{69.8 \small{$\pm 0.8$}} & 85.9 \small{$\pm 0.2$}          & 85.9 \small{$\pm 0.2$}          & 81.1 \small{$\pm 0.5$}          & 71.2 \small{$\pm 1.3$}          & 79.0          \\
Reduction            & Min                       & \textbf{84.7 \small{$\pm 0.1$}} & 63.3 \small{$\pm 0.7$}          & 85.6 \small{$\pm 0.2$}          & 88.4 \small{$\pm 0.9$}          & \textbf{85.0 \small{$\pm 0.6$}} & 78.6 \small{$\pm 0.8$}          & 80.9          \\
Rollout Mix          & Four-model mixture (16)   & 83.3 \small{$\pm\,0.3$}          & \textbf{69.2 \small{$\pm\,0.3$}} & \textbf{86.6 \small{$\pm\,0.2$}} & \textbf{89.9 \small{$\pm\,0.5$}} & 84.5 \small{$\pm\,0.1$}          & \textbf{77.8 \small{$\pm\,0.1$}} & \textbf{81.9} \\
Rollout Mix          & Two-model mixture (32)    & 83.3 \small{$\pm 0.1$}          & 67.4 \small{$\pm 0.2$}          & 86.0 \small{$\pm 0.0$}          & 89.4 \small{$\pm 0.0$}          & 84.0 \small{$\pm 0.5$}          & 76.5 \small{$\pm 1.0$}          & 81.1          \\
Rollout Mix          & Single-model (llama3-8B)  & \textbf{83.6 \small{$\pm 0.4$}} & 68.9 \small{$\pm 0.5$}          & 86.2 \small{$\pm 0.1$}          & 88.5 \small{$\pm 2.3$}          & 82.6 \small{$\pm 0.3$}          & 75.5 \small{$\pm 1.2$}          & 80.9          \\
Rollout Mix          & Single-model (qwen2.5 7B) & 82.9 \small{$\pm 0.7$}          & 65.7 \small{$\pm 0.5$}          & 85.6 \small{$\pm 0.2$}          & 88.6 \small{$\pm 2.0$}          & 84.3 \small{$\pm 0.2$}          & 75.2 \small{$\pm 0.2$}          & 80.4          \\
Supervision Schedule & D64 S\{1,2\}              & 83.3 \small{$\pm\,0.3$}          & 69.2 \small{$\pm\,0.3$}          & 86.6 \small{$\pm\,0.2$}          & 89.9 \small{$\pm\,0.5$}          & 84.5 \small{$\pm\,0.1$}          & 77.8 \small{$\pm\,0.1$}          & 81.9          \\
Supervision Schedule & D64 S4                    & \textbf{84.1}                   & 69.3                            & 86.5                            & \textbf{90.6}                   & 82.7                            & \textbf{78.9}                   & \textbf{82.0} \\
Supervision Schedule & D64 S8                    & \textbf{84.1}                   & 68.9                            & 86.5                            & 89.7                            & 84.2                            & 78.3                            & 81.9          \\
Supervision Schedule & D32 S4                    & 82.7                            & 69.2                            & 86.4                            & 89.1                            & 84.6                            & 77.4                            & 81.6          \\
Supervision Schedule & D32 S8                    & 83.9                            & 69.3                            & \textbf{86.8}                   & 90.2                            & 84.2                            & 77.8                            & \textbf{82.0} \\
Supervision Schedule & D16 S4                    & 84.3                            & 69.4                            & 86.7                            & 89.5                            & 84.2                            & 77.4                            & 81.9          \\
Supervision Schedule & D16 S8                    & 83.8                            & 69.7                            & 86.6                            & 89.5                            & 84.8                            & 77.6                            & \textbf{82.0} \\
Supervision Schedule & D1 S4                     & 82.7                            & \textbf{69.6}                   & 86.3                            & 88.1                            & 84.4                            & 75.4                            & 81.1          \\
Supervision Schedule & D1 S8                     & 83.1                            & 69.5                            & 86.6                            & 89.5                            & \textbf{85.8}                   & 77.1                            & 81.9          \\ \bottomrule
\end{tabular}%
}
\caption{Dataset-level output-moderation breakdown for the ablations in Table~\ref{tab:ablations}. This appendix table decomposes the aggregate output-moderation result from the main text into per-benchmark F1 scores under the same streaming evaluation protocol used throughout the paper, where a response is marked unsafe if the guardrail triggers at any point during generation. Read together with the main ablation table's streaming-timing and overblocking columns, these results support the central ablation story: \streamguard remains competitive across a broad range of target-construction variants, while the meaningful differences come from the operating-point trade-off between more aggressive early intervention and better benign calibration. Avg denotes the macro-average across benchmarks; supervision-schedule notation follows Table~\ref{tab:ablations}; reported values are mean $\pm$ standard deviation over three seeds where shown.}
\label{tab:ablations-output}
\end{table}

\section{Cross-Tokenizer Transfer Details}
\label{app:cross-tokenizer}

This appendix provides the dataset-level breakdown underlying the aggregated transfer results in Table~\ref{tab:cross-tokenizer-results}. We retain the same native \sgsmall reference and the same transferred Gemma and Qwen models discussed in Section~\ref{sec:results}. The two appendix tables separate a pattern that is compressed in the main transfer table: input-side transfer is somewhat uneven across benchmark families, whereas output-side transfer is substantially more consistent and accounts for most of the gain on streaming-related metrics.

\subsection{Input Moderation}
Table~\ref{tab:cross-tokenizer-input} reports per-benchmark input-moderation F1 for the native and transferred models. Input-side transfer is possible, but it is less uniform than output-side transfer.

The native \sgsmall reference remains strongest on the macro-average (86.4), driven by clear leads on Aegis (89.0), \aegisii (87.7), and WildGuard (88.6). The transferred \sgqwenmedium model comes closest at 85.8 and actually exceeds the native model on \sstest (98.8 vs.\ 98.6) and \harmbench (99.8 vs.\ 94.8), but it trails on Aegis, \aegisii, and WildGuard, which keeps its overall average below the native reference. \sggemmamedium reaches 85.1, with near-reference performance on \toxicchat, \oaimod, \aegisii, and WildGuard, but lower scores on Aegis and especially \harmbench.

The smaller transferred models illustrate the capacity limitation more clearly: \sggemmasmall falls to 79.9 on the macro-average, with especially weak \toxicchat/\oaimod results, which indicates that transfer alone does not compensate for limited model capacity. Overall, the input-side table supports the main-text claim that transfer is possible but less uniform on standard input moderation than on output-side or streaming behavior.
\begin{table}[]
\centering
\resizebox{\textwidth}{!}{%
\begin{tabular}{@{}llcccccccc@{}}
\toprule
model                                 & Cross-Tokenizer & ToxiC                  & OAIMod                 & Aegis                           & Aegis2                          & SSTest                          & HarmB                           & WildG                           & Avg           \\ \midrule
\qwensgsmallstrict                    & No              & 72.0                   & 68.3                   & 85.2                            & 84.9                            & 98.0                            & 97.2                            & 87.1                            & 84.7          \\
\qwensgsmallloose                     & No              & \textbf{75.5}          & \textbf{76.0}          & 77.7                            & 81.7                            & 96.9                            & 96.8                            & 86.0                            & 84.4          \\
\sgsmall \textbf{(ours)}              & No              & 74.9\small{$\pm\,0.3$} & 71.5\small{$\pm\,0.2$} & \textbf{89.0\small{$\pm\,0.3$}} & \textbf{87.7\small{$\pm\,0.2$}} & 98.6\small{$\pm\,0.2$}          & 94.8\small{$\pm\,2.9$}          & \textbf{88.6\small{$\pm\,0.2$}} & \textbf{86.4} \\
\sggemmasmall \textbf{(ours)}         & Yes             & 60.1 \small{$\pm 3.0$} & 63.8 \small{$\pm 0.6$} & 83.9 \small{$\pm 0.5$}          & 82.9 \small{$\pm 0.5$}          & 96.0 \small{$\pm 0.6$}          & 89.1 \small{$\pm 4.5$}          & 83.7 \small{$\pm 0.7$}          & 79.9          \\
\sggemmamedium \textbf{(ours)}        & Yes             & 74.0 \small{$\pm 0.4$} & 71.2 \small{$\pm 0.4$} & 87.4 \small{$\pm 1.3$}          & 86.6 \small{$\pm 0.2$}          & 98.5 \small{$\pm 0.0$}          & 89.5 \small{$\pm 2.8$}          & 88.4 \small{$\pm 0.2$}          & 85.1          \\
\sgqwensmall \textbf{(ours)}          & Yes             & 66.9 \small{$\pm 0.8$} & 67.7 \small{$\pm 0.8$} & 86.0 \small{$\pm 0.3$}          & 85.1 \small{$\pm 0.3$}          & 96.9 \small{$\pm 0.5$}          & 88.1 \small{$\pm 0.9$}          & 85.8 \small{$\pm 0.4$}          & 82.3          \\
\sgqwenmedium \textbf{(ours)}         & Yes             & 72.1 \small{$\pm 1.9$} & 71.1 \small{$\pm 0.6$} & 85.3 \small{$\pm 0.9$}          & 85.7 \small{$\pm 0.4$}          & \textbf{98.8 \small{$\pm 0.3$}} & \textbf{99.8 \small{$\pm 0.2$}} & 87.9 \small{$\pm 0.3$}          & 85.8          \\ \bottomrule
\end{tabular}%
}
\caption{Dataset-level input-moderation results for the cross-tokenizer transfer setting, complementing Table~\ref{tab:cross-tokenizer-results}. ``Cross-Tokenizer'' indicates whether prefix-level future-risk supervision is transferred from a source model with a different tokenizer than the target model. Despite the tokenizer mismatch, transferred \streamguard models remain competitive on standard input-moderation benchmarks. Avg denotes the macro-average across datasets; reported values for \streamguard models are mean $\pm$ standard deviation over three seeds where shown.}
\label{tab:cross-tokenizer-input}
\end{table}

\subsection{Output Moderation}
Table~\ref{tab:cross-tokenizer-output} reports per-benchmark output-moderation F1 under the same streaming evaluation protocol used throughout the paper. The output-side pattern is more consistent.

\sggemmamedium achieves the best transferred macro-average (81.3), driven by the best \beavertails (87.2), \xstest (87.2), and \aegisii (84.7) scores together with a strong \srlhf result (69.7). In the main transfer table, it also gives the best transferred streaming F1 (98.2) and miss rate (3.5\%), although its OnTime rate (92.3\%) is slightly below the native \sgsmall reference (92.9\%). The native \sgsmall reference remains best on WildGuard (77.2), but is otherwise matched or exceeded by the stronger transferred backbones. The transferred Qwen models are also competitive: both Qwen variants reach 68.8 on \srlhf and remain close to the native model on \harmbench and \beavertails, though they are weaker on \xstest and WildGuard than \sggemmamedium.

The small \sggemmasmall model again shows that transfer still depends on model capacity: it retains reasonable \beavertails performance (85.2) but drops sharply on \xstest (76.1) and WildGuard (67.1), pulling its average down to 75.3. These finer-grained results support the main practical takeaway: future-risk supervision transfers well across tokenizer families for output moderation, but the strength of the transfer still depends on target-model capacity and family-specific calibration.
\begin{table}[]
\centering
\resizebox{\textwidth}{!}{%
\begin{tabular}{@{}lllllllll@{}}
\toprule
model                                    & Cross-Tokenizer & HarmB                  & SRLHF                           & BeaverTails                     & \xstestr                         & Aegis2                          & WildG                           & Avg           \\ \midrule
\qwensgsmallstrict                       & No              & \textbf{83.1}          & 62.8                            & 84.5                            & 84.8                            & 81.4                            & 76.3                            & 78.8          \\
\qwensgsmallloose                        & No              & 80.6                   & 61.7                            & 84.0                            & 83.3                            & 81.4                            & 75.8                            & 77.8          \\
\sgsmall \textbf{(ours)}                 & No              & 82.4 \small{$\pm\,0.2$} & 68.0 \small{$\pm\,0.7$}          & 86.2 \small{$\pm\,0.1$}          & 84.6 \small{$\pm\,0.7$}          & 82.9 \small{$\pm\,0.3$}          & \textbf{77.2 \small{$\pm\,0.3$}} & 80.2          \\
\sggemmasmall (\textbf{ours})            & Yes             & 77.2 \small{$\pm 2.8$} & 65.6 \small{$\pm 0.4$}          & 85.2 \small{$\pm 0.4$}          & 76.1 \small{$\pm 2.4$}          & 80.7 \small{$\pm 0.6$}          & 67.1 \small{$\pm 3.4$}          & 75.3          \\
\sggemmamedium (\textbf{ours})           & Yes             & 82.7 \small{$\pm 0.8$} & 69.7 \small{$\pm 0.4$}          & \textbf{87.2 \small{$\pm 0.2$}} & \textbf{87.2 \small{$\pm 0.0$}} & \textbf{84.7 \small{$\pm 0.3$}} & 76.3 \small{$\pm 1.0$}          & \textbf{81.3} \\
\sgqwensmall (\textbf{ours})             & Yes             & 80.9 \small{$\pm 0.9$} & \textbf{68.8 \small{$\pm 0.4$}} & 86.3 \small{$\pm 0.4$}          & 82.8 \small{$\pm 0.8$}          & 83.2 \small{$\pm 0.7$}          & 74.4 \small{$\pm 0.8$}          & 79.4          \\
\sgqwenmedium (\textbf{ours})            & Yes             & 82.8 \small{$\pm 1.1$} & \textbf{68.8 \small{$\pm 1.3$}} & 86.8 \small{$\pm 0.2$}          & 83.0 \small{$\pm 1.7$}          & 83.1 \small{$\pm 0.4$}          & 76.0 \small{$\pm 1.5$}          & 80.1          \\ \bottomrule
\end{tabular}%
}
\caption{Dataset-level output-moderation results for the cross-tokenizer transfer setting, complementing Table~\ref{tab:cross-tokenizer-results}. Models are evaluated under the same streaming output-moderation protocol used in the main paper, where a response is marked unsafe if it triggers at any point during generation. Cross-tokenizer transfer remains effective across target families, with transferred \streamguard models preserving strong output-moderation performance. Avg denotes the macro-average across datasets; reported values for \streamguard models are mean $\pm$ standard deviation over three seeds where shown.}
\label{tab:cross-tokenizer-output}
\end{table}

\begin{table}[]
\centering
\begin{tabular}{@{}llcccc@{}}
\toprule
\multirow{2}{*}{Model}  & \multirow{2}{*}{Role} & \multirow{2}{*}{Latency (ms)} & \multicolumn{2}{c}{Decision Ratio} & \multirow{2}{*}{Decision/s} \\ \cmidrule(lr){4-5}
                        &                       &                               & 8B               & 70B             &                             \\ \midrule
Llama3-8B-Instruct      & Generator             & 9.8                           & -                & -               & -                           \\
Llama3-70B-Instruct     & Generator             & 99.2                          & -                & -               & -                           \\ \cmidrule(lr){1-6}
\sggemmasmall          & Guardrail             & 2.4                           & 0.243            & 0.024           & 420                         \\
\sggemmamedium         & Guardrail             & 3.7                           & 0.375            & 0.037           & 271                         \\
\sgqwensmall           & Guardrail             & 4.3                           & 0.439            & 0.043           & 232                         \\
\sgqwenmedium          & Guardrail             & 4.9                           & 0.501            & 0.050           & 203                         \\
\sgsmall               & Guardrail             & 3.2                           & 0.330            & 0.033           & 309                         \\
\sgmedium              & Guardrail             & 6.0                           & 0.616            & 0.061           & 165                         \\
\sglarge               & Guardrail             & 9.5                           & 0.969            & 0.096           & 105                         \\ \bottomrule
\end{tabular}%
\caption{
Latency measurements in the steady-state decoding regime. We measure from a prefix length of 1024 tokens and average over the next 1024 decoding steps, using 100 runs per setup. All experiments use NVIDIA H100 GPUs with HuggingFace and \texttt{StaticCache}; Llama3-70B-Instruct is measured on two GPUs. Decision ratios normalize guardrail latency by the average per-token latency of the Llama3-8B-Instruct and Llama3-70B-Instruct generators and are the primary quantity of interest for pipelined moderation.
}
\label{tab:latency}
\end{table}

\section{Latency Measurements}
\label{app:latency}

This appendix reports latency measurements for streaming moderation. Our main setting is \emph{pipelined} moderation, in which the guardrail runs concurrently with decoding. In this regime, the key systems question is whether guardrail decisions can keep pace with the generator's token cadence, and therefore whether risky output can be blocked before additional tokens are served. For completeness, we also distinguish this setting from a simpler \emph{buffered} setup, but our interpretation and conclusions focus on the pipelined regime.

All measurements use NVIDIA H100 GPUs with the HuggingFace transformers library generation stack with StaticCache~\cite{wolf2020huggingfacestransformersstateoftheartnatural}. The Llama3-70B-Instruct generator is measured on two GPUs; all other models are measured on a single GPU. We measure latency in a steady-state regime by starting from a prefix of 1024 tokens and timing the next 1024 generated tokens. All reported results are averaged over 100 runs. We use Llama3-8B-Instruct and Llama3-70B-Instruct as generators.

For each generator, we report the average latency per emitted token, denoted \(\bar{g}\). For each guardrail, we report the average decision latency \(r\) in milliseconds and the corresponding throughput in decisions per second. To summarize pipelined performance, we additionally report the \emph{decision ratio}
\[
\rho = \frac{r}{\bar{g}},
\]
computed separately against each generator. This ratio normalizes guardrail latency by the generator's average per-token latency and directly captures whether moderation keeps pace with decoding.

In the pipelined setting, \(\rho\) is the main quantity of interest. Absolute latency remains useful because it exposes raw runtime directly, but the deployment implication depends on latency relative to the generator rather than latency in isolation. A guardrail with a few milliseconds of latency may be comfortably faster than one generator and much closer to the limit for another. The decision ratio captures this dependence directly.

The decision ratio also determines post-decision exposure in the pipelined regime. Under an average-cadence approximation, the additional number of tokens served after a block decision becomes available is
\[
L_{\mathrm{extra}} = \max\left(0, \left\lceil \rho \right\rceil - 1\right)
= \max\left(0, \left\lceil \frac{r}{\bar{g}} \right\rceil - 1\right),
\]
with the smoother approximation \(L_{\mathrm{extra}} \approx \max(0, \rho - 1)\). In particular, when \(\rho < 1\), the guardrail decision arrives before the next token would be served, so the next token does not reach the user. When \(1 < \rho \le 2\), roughly one additional token may be served before intervention, and larger values correspond to proportionally larger exposure. Thus, \(\rho\) summarizes both throughput compatibility and post-decision exposure for pipelined moderation.

For completeness, we also distinguish pipelined execution from a buffered setup. In a buffered setup, generated tokens are briefly held before display, so the generator does not pause for every guardrail decision. The resulting latency is therefore not a strict per-token sum of generation and guardrail time. If the generator is faster than the guardrail, buffering mainly affects the initial release of output, after which visible tokens are emitted at approximately the guardrail's rate. If the guardrail is faster than the generator, the two stages overlap and the added delay is correspondingly smaller. We include this setup only as a reference point; the main analysis in this section concerns pipelined moderation.

Table~\ref{tab:latency} reports the measured results. Llama3-8B-Instruct runs at 9.8\,ms/token, while Llama3-70B-Instruct runs at 99.2\,ms/token. Across the evaluated guardrails, decision latency ranges from 2.4\,ms to 9.5\,ms, corresponding to 105--420 decisions/s. Relative to the 8B generator, decision ratios range from 0.243 to 0.969. Relative to the 70B generator, they range from 0.024 to 0.096. Thus, for both generators, all evaluated guardrails satisfy \(\rho < 1\). Under the average-cadence interpretation above, this means that once a block decision is available, the next token is not served to the user. The tightest pairing is Llama3-StreamGuard-8B with Llama3-8B-Instruct at \(\rho = 0.969\), which is close to parity but still remains below this threshold. All other pairings provide additional latency headroom, especially for the 70B generator.

These results support the practical viability of pipelined streaming moderation. Even for the faster 8B generator, all guardrails keep pace on average, indicating that moderation can run ahead of exposure without becoming the throughput bottleneck in the measured regime. For the 70B generator, the margin is substantially larger, making pipelined deployment even more favorable.

Finally, the absolute latencies reported here should be viewed as measurements under a simple and reproducible baseline stack rather than as the lowest achievable deployment numbers. All experiments use HuggingFace with \texttt{StaticCache}. In practical serving deployments, specialized systems such as \texttt{vLLM} and optimized kernels can reduce both generator and guardrail latency further. We therefore expect lower absolute latencies in production settings, although the decision ratio remains the more relevant quantity for assessing whether a guardrail keeps pace with generation.

\end{document}